\documentclass[journal]{IEEEtran}
\usepackage{amsmath,amsfonts}
\usepackage{algorithmic}
\usepackage{array}
\usepackage[caption=false,font=scriptsize]{subfig}

\usepackage{textcomp}
\usepackage{stfloats}
\usepackage{url}
\usepackage{cite}
\usepackage{verbatim}
\usepackage{graphicx}
\usepackage{hyperref}
\usepackage{multirow}
\usepackage{textgreek, stfloats, booktabs}
\usepackage{diagbox}
\usepackage{makecell}

\usepackage{mathtools}

\usepackage{bbm}

\def\BibTeX{{\rm B\kern-.05em{\sc i\kern-.025em b}\kern-.08em
    T\kern-.1667em\lower.7ex\hbox{E}\kern-.125emX}}
\usepackage{balance}
\begin{document}
\title{Sample-Efficient Unsupervised Domain Adaptation of Speech Recognition Systems:\\ A case study for Modern Greek}
\author{Georgios Paraskevopoulos
\IEEEmembership{Student Member, IEEE}, Theodoros Kouzelis, Georgios Rouvalis, Athanasios Katsamanis \IEEEmembership{Member, IEEE}, Vassilis Katsouros \IEEEmembership{Member, IEEE}, Alexandros Potamianos \IEEEmembership{Fellow, IEEE}
\thanks{G. Paraskevopoulos is with the Graduate School
of ECE, National Technical University of Athens, Athens, Greece}
\thanks{G. Paraskevopoulos, T. Kouzelis, G. Rouvalis, A. Katsamanis, V. Katsouros are with the Institute for Speech and Language Processing, Athena Research Center, Athens, Greece}
\thanks{A. Potamianos is with the Faculty of ECE, National Technical University of Athens, Athens, Greece}
}


\maketitle

\begin{abstract}

Modern speech recognition systems exhibits rapid performance degradation under domain shift. This issue is especially prevalent in data-scarce settings, such as low-resource languages, where diversity of training data is limited.
In this work we propose M2DS2, a simple and sample-efficient finetuning strategy for large pretrained speech models, based on mixed source and target domain self-supervision. We find that including source domain self-supervision stabilizes training and avoids mode collapse of the latent representations. For evaluation, we collect HParl, a $120$ hour speech corpus for Greek, consisting of plenary sessions in the Greek Parliament. We merge HParl with two popular Greek corpora to create GREC-MD, a test-bed for multi-domain evaluation of Greek ASR systems. In our experiments we find that, while other Unsupervised Domain Adaptation baselines fail in this resource-constrained environment, M2DS2 yields significant improvements for cross-domain adaptation, even when a only a few hours of in-domain audio are available. When we relax the problem in a weakly supervised setting, we find that independent adaptation for audio using M2DS2 and language using simple LM augmentation techniques is particularly effective, yielding word error rates comparable to the fully supervised baselines.

\end{abstract}


\section{Introduction}
\label{sec:intro}

Automatic Speech recognition (ASR) models have matured to the point where they can enable commercial, real-world applications, e.g., voice assistants, dictation systems, etc., thus being one of machine learning's success stories.
However, the performance of ASR systems rapidly deteriorates when the test data domain differs significantly from the training data.
Domain mismatches can be caused by differences in the recording conditions, such as environmental noise, room reverberation, speaker and accent variability, or shifts in the target vocabulary.
These issues are extenuated in the case of low-resource languages, where diversity in the training data is limited due to poor availability of high-quality transcribed audio. Therefore, specialized domain adaptation approaches need to be employed when operating under domain-shift.

Unsupervised Domain Adaptation (UDA) methods are of special interest, as they do not rely on expensive annotation of domain-specific data for supervised in-domain training.
In contrast to supervised approaches, where the existence of labeled  data would allow to train domain-specific models, UDA methods aim to leverage data in the absense of labels to improve system performance in the domain of interest \cite{long2015learning,ganin2016domain}.
In the context of speech recognition the importance of UDA is extenuated, as the transcription and alignment process is especially expensive and time-consuming.
Adaptation methods have been explored since the early days of ASR, at different levels of the system and  different deployment settings \cite{bell2020adaptation}.
UDA has been used  to improve the robustness of ASR on a variety of recording conditions including far-field speech, environmental noise and reverberation \cite{selftrainASRSoftFirst,selftrainSoftpovey}, \cite{adverserialASR1}. Furthermore, UDA has been used for speaker adaptation, and to improve performance under speaker, gender and accent variability \cite{adverserialASRspeaker,adverserialASRaccent}.
UDA has also been employed for multilingual and cross-lingual ASR, in order to improve ASR models for low-resource languages \cite{adverserialASRlow}, adapt to different dialects \cite{asami2017domain}, and even train speech recognition systems for endangered languages \cite{NOWAKOWSKI2023103149}.

Classical speech adaptation techniques involve feature-based techniques, e.g.,
speaker normalization \cite{furui1980training}, feature-based approaches \cite{miao2014towards,parthasarathi2015fmllr,ivectors}, or multi-condition training \cite{hirsch2000aurora}.
Generally, traditional approaches require some knowledge about the target domain, and the domain mismatch, e.g., regarding the noise and reverberation variability \cite{7472774}, and require specific engineering for each adaptation scenario.

Modern ASR pipelines, increasingly rely on end-to-end neural networks, e.g., \cite{las,rnnt}, or large pretrained models with self-supervised objectives \cite{baevski2020wav2vec,conneau21_interspeech}. The key approaches employed for UDA of end-to-end ASR models can be grouped
in three categories, namely, teacher-student learning \cite{asami2017domain}, domain adversarial training
 \cite{denisov2018unsupervised}, and target domain self-supervision  \cite{google21domain}. The benefit of these techniques is that they do not require any special knowledge about the source or the target domain. This makes end-to-end UDA approaches versatile and able to be utilized in a larger array of adaptation scenarios. In particular, adaptation through self-supervision has been shown to be a robust, simple and efficient technique for adaptation of state-of-the-art speech models \cite{hsu21_interspeech}.

\begin{table*}[!t]
\caption{\label{tab:rw}Summary of related works on Unsupervised Domain Adaptation for ASR.}
\centering
\begin{tabular}{ccccc}
\hline
\textbf{Work} & \textbf{Method} & \textbf{Model}  & \textbf{Adaptation Setting} &  \multicolumn{1}{c}{\textbf{Language}}  \\ \hline
  \cite{google21domain, selftrainASRDUST, selftrainSoft1} & \makecell{Teacher-Student   \\ Hard and soft labels} & \makecell{Conformer RNN-T \cite{gulati2020conformer} \\ Transformer CTC \\ RNN-T \cite{rnnt} }& \makecell{News speech, Voice search, Far-field, \\
Telephony, YouTube} & \makecell{English} \\ \hline
\cite{selftrainASRSoftFirst, selftrainSoftpovey} & \makecell{Teacher-Student \\ Soft labels}  & \makecell{TDNN-LSTM\cite{lstmrnn}}& \makecell{Noise, Far-field} & \makecell{English} \\ \hline
 \cite{selftrainSoftandHard} & \makecell{Teacher-Student \\ Hard and soft labels }  &  NiN-CNN \cite{nin}  &\makecell{Dialects \\ Children speech} & \makecell{Japanese} \\ \hline
 \cite{selftrainSoft2} & \makecell{Teacher-Student \\ Soft labels }  &Streaming RNN-T \cite{streamingrnnt}&\makecell{Multilingual} & \makecell{English, \\Brazilian Portuguese, \\Russian, \\, Turkish, \\ Nordic/Germanic} \\ \hline


\cite{adverserialASR1, adverserialASR2, adverserialASR3} & \makecell{Domain Adversarial Training}  &\makecell{TDNN Kaldi \cite{peddinti2015time,Povey_ASRU2011} \\ DNN-HMM \\ DNN-HMM }& \makecell{Noise, Channel} & \makecell{English} \\ \hline
\cite{adversarialASRFarfield} & \makecell{Domain Adversarial Training } &RNN-CTC \cite{rnnctc} & \makecell{Far-field} & \makecell{English} \\ \hline
\cite{adverserialASRaccent, adversarialASRAccentREDAT} & \makecell{Domain Adversarial Training} &\makecell{TDNN Kaldi  \\ RNN-T  } & \makecell{Accent} & \makecell{Mandarin} \\ \hline
\cite{adverserialASRspeaker, adversarialASRraw} & \makecell{Domain Adversarial Training}  &\makecell{DNN-HMM \\ CNN-DNN } & \makecell{Speaker, Gender, \\ Accent} & \makecell{English} \\ \hline
\cite{adverserialASRlow} & \makecell{Domain Adversarial Training }  & DSN \cite{DSN} &\makecell{Multilingual} & \makecell{Hindi, Sanskri} \\ \hline

\cite{hsu21_interspeech, zhu2022boosting} & \makecell{Continual Pre-Training }  & wav2vec2 \cite{baevski2020wav2vec}&\makecell{Audiobooks, Accents, \\ Ted Talks,  Telephony, \\ Crowd-sourced, Parlamentary speech} & \makecell{English} \\ \hline
\cite{kim22e_interspeech} & \makecell{Continual Pre-Training } & wav2vec2 & \makecell{Cross-lingual} & \makecell{Korean} \\ \hline
\cite{dehaven2022improving,NOWAKOWSKI2023103149} & \makecell{Continual Pre-Training } & \makecell{XLSR-53 \cite{conneau21_interspeech} \\ wav2vec2}& \makecell{Low resource languages} & \makecell{Ainu \\ Georgian, Somali, \\ Tagalog, Farsi} \\ \hline

\end{tabular}
\end{table*}

Here, we leverage in-domain self-supervision to propose the Mixed Multi-Domain Self-Supervision (M2DS2) finetuning strategy, enabling sample-efficient domain adaptation of wav2vec2 \cite{baevski2020wav2vec} based speech recognition models, even when available in-domain data are scarce. Our key contributions are organized as follows:

\begin{enumerate}
    \item Inspired by recent advances on UDA for Natural Language Processing systems \cite{karouzos-etal-2021-udalm}, we propose a finetuning strategy for speech models, where the self-supervised objective is based on a contrastive loss in Section~\ref{sec:method}. Contrary to prior works, who leverage only in-domain self-supervision, we find that in this contrastive setting this leads to mode-collapse of the latent representations, and mixed source and target domain self-supervision is essential. We demonstrate this empirically in Section~\ref{sec:multi-domain}.
    \item We collect and curate HParl, the largest publicly available\footnote{We plan to release this version of HParl under the CC BY-NC 4.0 license upon publication. The other corpora used in this work are available through their respective distributors.} speech corpus for Greek, collected from plenary sessions in the Greek Parliament between 2018 and 2022. We establish a data collection, pre-processing and alignment pipeline that can be used for continuous data integration, as the parliamentary proceedings get regularly uploaded. We provide a detailed description of our data collection process and the dataset statistics in Section~\ref{sec:hparl-collection}. HParl is merged in Section~\ref{sec:grec-md-corpus} with two popular Greek corpora (Logotypografia and CommonVoice) to create GREC-MD, a testbed for multi-domain evaluation of ASR systems in Greek.
    \item We demonstrate that, while other baselines fail at UDA in our resource-constrained setting, M2DS2 can improve model performance in the target domain in multiple adaptation scenarios in Section~\ref{sec:experiments}. Specifical emphasis is given in the sample efficiency of our approach in Section~\ref{sec:sample-eff}, where we demonstrate successful adaptation even when we reduce the available in-domain data.
    \item When we relax the problem to a weakly supervised adaptation setting, where some in-domain text is available but the pairing between audio and text is unknown, we find that M2DS2 can be effectively combined with simple N-gram adaptation techniques to get comparable performance with the fully supervised baseline in Section~\ref{sec:weak}. Furthermore we find that a simple text augmentation approach, based on perplexity filtering of a large corpus can produce strong adaptation results, even for small amounts of in-domain text.
\end{enumerate}
Additionally, we provide a formulation of the UDA problem for ASR in Section~\ref{sec:problem-def} and link prior works to this formulation in Sections~\ref{sec:teacher},~\ref{sec:dat} and~\ref{sec:self-sup-background}. We provide detailed experimental settings for reproducibility in Section~\ref{sec:settings}, and an upper-bound estimation for UDA performance with fully supervised finetuning in Section~\ref{sec:supervised}.

\section{Background}
\label{sec:background}
We start by formally defining the Unsupervised Domain Adaptation (UDA) problem. Initially, we formulate the problem in a classification setting and then we extend it for speech recognition. We then provide an overview of different adaptation approaches in the literature, and link each approach to the UDA problem formulation. Table~\ref{tab:rw} presents a summary of the key adaptation settings and applications that are explored in the literature. We see, that a relatively small  amount of methods, and their variants, is used to address multiple real-world ASR problems, for example, cross-lingual, accent, speaker and noise adaptation. Furthermore, while the majority of the works focus on the English language, there is an effort to explore other popular languages, e.g., Mandarin, and under-resourced languages, e.g., Ainu, Somali etc.

\subsection{Problem Definition}
\label{sec:problem-def}

Formally, the problem of UDA can be defined as follows.
Let $X \subseteq \mathbb{R}^n$ be a real-valued space that consists of $n$-dimentional feature vectors $x \in X$, and $Y$ a finite set of labels $y\in Y$, i.e., $Y = \{1, 2, \dots, L\}$.
Furthermore, assume two different distributions, i.e., the source domain distribution $\mathcal{S}(x,y)$ and the target domain distribution $\mathcal{T}(x,y)$,  defined  on the cartesian product $X \times Y$.

The goal is to train a model that learns a mapping between feature vectors $x_{\mathcal{T}}$ to their respective labels $y_{\mathcal{T}}$ for samples drawn from the target distribution $(x_{\mathcal{T}},y_{\mathcal{T}}) \sim \mathcal{T}$.

At training time we have access to samples from the source distribution $\mathcal{S}(x, y)$ and the marginalized target distribution $\mathcal{T}(x)$, i.e., no target labels are provided. We define the training dataset $D$ as the concatenation of the source and target training sets, $D = (D_S, D_T)$. $D_S$ and $D_T$ are defined as sequences of tuples, i.e.,

\begin{equation}\label{eq:train-sets}
\begin{split}
 &	D_S = \{(x_i, y_i) \, |\, (x_i, y_i) \sim \mathcal{S}(x, y) ,\, 1 \leq i \leq N \} \\
 &  D_T =\{(x_i, \emptyset) \, | \, x_i \sim \mathcal{T}(x) , \, 1 \leq i \leq M \},
\end{split}
\end{equation}

where we draw $N$ samples from $\mathcal{S}(x,y)$ and $M$ samples from $\mathcal{T}(x)$. Finally, we augment tuples in $D$ with a domain indicator function:

\begin{equation}
    \begin{split}
    D &= \{(x_i, y'_i, \mathbbm{1}_i) \, | \, 1 \leq i \leq N+M \}\\
    \mathbbm{1}_i &= \begin{cases}
0 ~&\text{ if }~ x_i \sim \mathcal{S}(x), \\
1 ~&\text{ if }~ x_i \sim \mathcal{T}(x).
\end{cases}\\
    y'_i &= \begin{cases}
y_i ~&\text{ if }~ x_i \sim \mathcal{S}(x), \\
\emptyset ~&\text{ if }~ x_i \sim \mathcal{T}(x).
\end{cases}
    \end{split}
\end{equation}

\subsubsection{Unsupervised (Acoustic) Adaptation for ASR}
The above definition can be directly extended in the case of speech recognition, with some modifications. In detail, we modify the feature space $X$, to be the set of (finite) sequences of real-valued feature vectors $(x_k)_{k\in \mathbb{N} \setminus \{\infty\}} \in X \subseteq (\mathbb{R}^n)^*$. Furthermore, the label space $Y$ is modified to be the set of sequences $(y_m)_{m\in \mathbb{N} \setminus \{\infty\}}$, where $Y = (\{1, 2,\dots, L\})^*$ contains finite-length sequences over a finite lexicon. For CTC training we make the assumption that $k>m$ for any sample $(x_k, y_m)$, i.e., feature sequences are longer than their respective label sequences \cite{10.1145/1143844.1143891}.
The rest of the definitions need no modifications.

\subsubsection{Unsupervised (Language) Adaptation for ASR}
\label{sec:unsup-lang-prob}
Adaptation for ASR systems can also be performed at the language level, i.e., the label space. In this setting, we assume that the target domain samples are drawn from the marginalized target distribution $\mathcal{T}(y)$. The target dataset $D_T$ now consists of tuples in the form $(\emptyset, y_i)$, where $y_i$ is the label word sequence $(y_m)_{m\in \mathbb{N} \setminus \{\infty\}}$ for the $i$-th sample.

\subsubsection{Weakly supervised Adaptation for ASR}
\label{sec:weak-lang-prob}
The last setting we explore is the case were both audio and language in-domain samples are available, but the mapping between them is unknown. This situation can be encountered in real-world settings, e.g., in the case in-domain audio and text are collected independently. For example consider the case where audio clips from news casts are collected, along with contemporary newspaper articles. Another example is the case where long audio clips alongside with transcriptions are available, but no fine-grained time alignments\footnote{While a fully supervised in-domain dataset can be constructed in this case using long / forced alignment methods, this is not a focal point for the experimental part of this work.}.
In this case the target domain samples are drawn independently from the marginalized distributions  $\mathcal{T}(x)$ and  $\mathcal{T}(y)$, and the target dataset $D_T$ consists of tuples in the form $(x_i, \emptyset)$ and $(\emptyset, y_i)$.

\subsection{Teacher-Student Models}
\label{sec:teacher}

Teacher-Student learning or self-training, is one of the earliest methods in semi-supervised learning \cite{selftrain1,selftrain2,selftrain3}.
The key idea is to reduce the problem of unsupervised learning of the task at hand in the target domain to a supervised one. The general methodology is to train a teacher model $g_S$ using the labeled data in the source domain $D_S$, and then use this for inference on the target domain to produce pseudolabels $\hat{y}_i = g_S(x_i), \, x_i \sim \mathcal{T}(x)$. The target domain dataset $D_T$ is augmented with these silver labels, to contain tuples $(x_i, \hat{y}_i)$. Finally, a student model $g_T$ is trained in a supervised fashion, using the augmented $D_T$ or a combination of $D_S$ and $D_T$. This process is usually repeated, with the student model serving as the teacher model for the next iteration, until no further improvement is observed.
More recently, soft target
Teacher-Student learning has been explored for ASR \cite{selftrainSoft1, selftrainSoft2, sleftrainSoft3}, where the KL divergence between the teacher and student output label distributions
is used as the loss function.

Being trained only on the source domain data the teacher model is susceptible to error propagation.
Filtering is a commonly used technique to achieve the right
balance between the size of the target domain used for training the student model and the noise in the pseudolabels.
Confidence scoring based on the likelihood is usually applied, discarding those utterances for which the hypothesized labels are untrustworthy  \cite{selftrainASR4}.
In \cite{selftrainASRDUST} dropout is used to measure the model uncertainty. The agreement between model predictions with and without dropout are used for confidence scoring.
 In \cite{google21domain} a  multi-task training objective with a confidence loss is applied to minimise
the binary cross entropy between the estimated confidence
and the binary target sequence.
In order to learn more robust and generalizable features from the teacher model,  Noisy Student Training (NST) has been proposed in \cite{Park_2020}.
The teacher models generates pseudolabels for $D_T$ while the student models are trained on a heavily augmented version of $D_T$ \cite{Park_2020}. In \cite{Park_2020, nstNeurips} the augmentation of the input target data is performed with SpecAugment \cite{park2019specaugment}, while in \cite{selftrainSoftandHard} a spectrum frequency augmentation is performed.

In \cite{selftrainASRSoftFirst} Teacher-Student learning with soft labels is introduced for ASR to tackle noisy, far-field, and children speech. In \cite{selftrainSoftpovey}, this approach is extended for LF-MMI based models and used for noisy, far-field and bandwidth  adaptation. In  \cite{selftrainSoftandHard} a weighted sum of hard and soft target cross entropy losses is used for Japanese dialects and children speech adaptation.  Ramabhadran et al. \cite{selftrainSoft2} propose a self-adaptive distillation, and a method for distilling from
multiple teachers that is applied across several multilingual ASR
systems for different language groups.
A comparison between soft and hard targets for RNN-T models \cite{rnnt} showed that soft targets perform better
when both the teacher and student models have the same architecture. Otherwise, hard targets are superior \cite{sleftrainSoft3}.

\subsection{Domain Adversarial Training}
\label{sec:dat}

Domain Adversarial Training (DAT) was initially introduced for image classification \cite{imageadverserial}. The key idea is to train a model that learns deep features that solve the task at hand in the source domain, while being invariant with respect to the domain shift. Concretely, the model is trained end-to-end using a combination of the supervised task loss $L_t$, learned on $D_S$, and the domain discrimination loss $L_{a}$, i.e., $L = L_t - \alpha L_{a}$. The loss $L_a$ is binary cross-entropy, trained for domain discrimination using the tuples $(x_i, \mathbbm{1}_i)$. Notice the $-$ sign in the loss indicates adversarial learning, i.e., the model should learn features that cannot discriminate between domains, while solving the task.

In \cite{adverserialASR1} DAT is employed for noise adaptation on a noise corrupted version of WSJ \cite{paul1992design} as the target dataset. Using the Aurora-4 \cite{yeung2004improved}  dataset which has labels associated to the noise type, Serdyuk et al. \cite{adverserialASR2} train an adversarial noise classifier.  In \cite{adverserialASRaccent} and \cite{adversarialASRAccentREDAT} DAT is utilized for accent adaptation for Mandarin and English respectively.
Anoop C.S. et al. \cite{adverserialASRlow}  propose DAT, to address the
scarcity of data in low-resource languages which share a
common acoustic space with a high-resource language, namely Sanskrit and Hindi. They empirically demonstrate the effectiveness of adversarial training, presenting experiments with and without the reversal of the domain classification loss.

\subsection{Leveraging In-domain Self-supervision}
\label{sec:self-sup-background}
These lines of work have roots in Natural Language Processing tasks \cite{karouzos-etal-2021-udalm,gururangan-etal-2020-dont}, and explore domain adaptation by leveraging the in-domain data $D_T$ for self-supervised learning. The core focus is domain adaptation of large pre-trained models, e.g., \cite{bert}, and self-supervision is achieved by use of the pre-training self-supervised loss $L_s$. This process can either take part in stages, via continual pre-training \cite{gururangan-etal-2020-dont}, or by constructing a multitask objective $L = L_t + \alpha L_s$, as in \cite{karouzos-etal-2021-udalm}.

Continual Pre-Training (CPT) has been explored for adaptation of ASR models.
Robust wav2vec2 \cite{hsu21_interspeech} explores the effectiveness of CPT for domain adaptation, indicating the importance of utilizing unlabeled in-domain data.
In CASTLE \cite{zhu2022boosting}, CPT is combined with an online pseudolabeling strategy for domain adaptation of wav2vec2. Cross-dataset evaluation for popular English speech corpora indicates that CPT helps to reduce the error rate in the target domain.
In \cite{kim22e_interspeech} and \cite{NOWAKOWSKI2023103149}
CPT is utilized for cross-lingual adaptation of wav2vec2 for Korean and Ainu respectively. Notably for Ainu, which is an endagered language, CPT has resulted in significant system improvement. DeHaven and Jayadev \cite{dehaven2022improving} compare CPT and pseudolabeling for adapting XLSR-53 to four  under-resourced languages, i.e., Georgian, Somali, Tagalog and Farsi. They find that both approaches yield similar improvements, with CPT being the more computationally efficient approach.

While CPT yields significant improvements in a variety of tasks, one common theme in these works is the assumption of hundreds or thousands of hours of available in-domain data, mostly from online resources, e.g., YouTube. This can be infeasible when we consider more niche adaptation settings, or possible privacy concerns, e.g., how would one collect $1000$ hours of psychotherapy sessions in Greek? In this work, we explore domain adaptation methods in a more resource-constrained environment.

\begin{figure}[t!]
    \centering
    \includegraphics[width=\columnwidth]{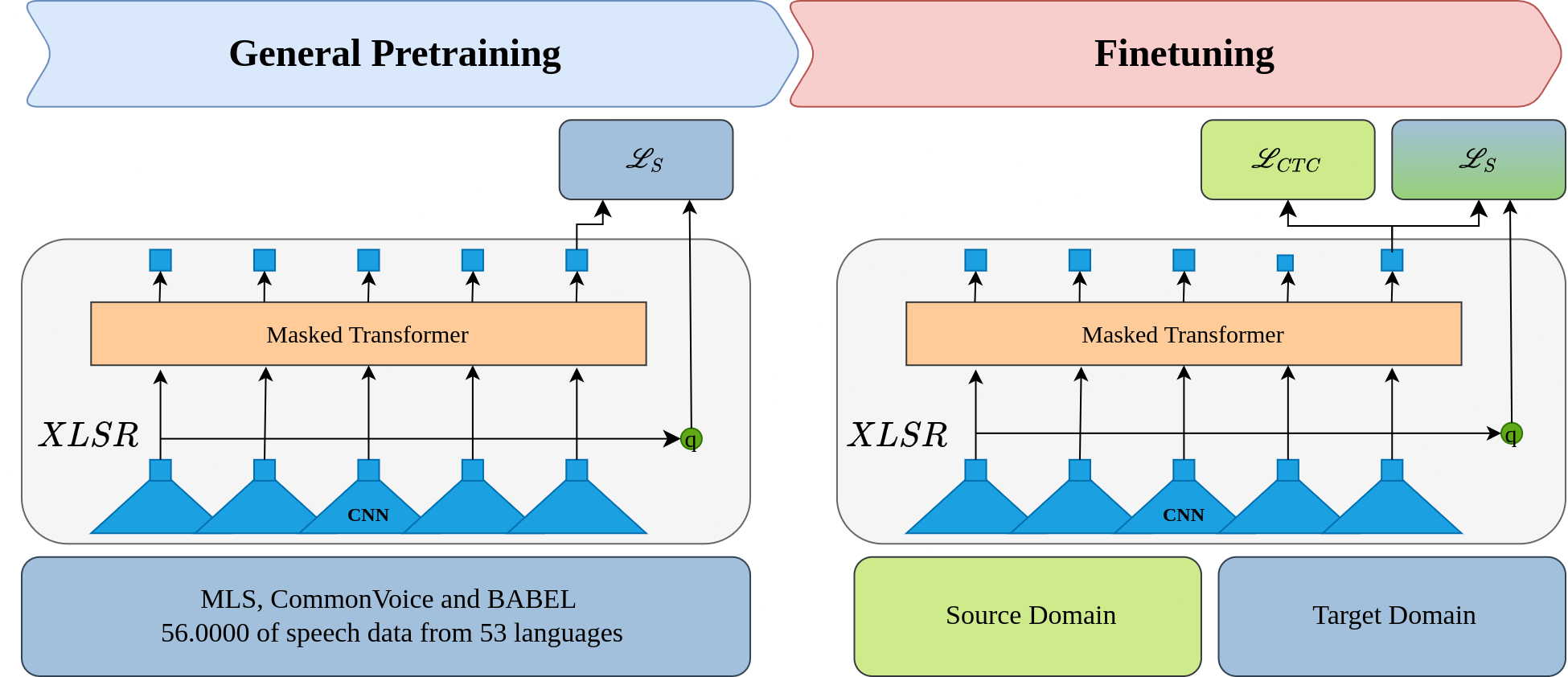}
    \vspace{5mm}
    \caption{Target-domain adaptation through self-supervision. In the left we see the general pre-training stage of XLSR-53 using the self-supervised loss $L_s$. General pre-training is performed on $56,000$ hours of audio in $53$ languages. In the right, we see the proposed domain-adaptive finetuning stage, where the speech recognition task is learned using transcribed source domain data, while adaptation to the target domain is performed by including the self-supervised loss over (audio-only) source and target domain data}
    \label{fig:m2ds2}
\end{figure}

\section{Domain Adaptation Through Multi-Domain Self-Supervision}
\label{sec:method}

The proposed approach is based on end-to-end adaptation of a large pre-trained speech model during the finetuning phase, by including in-domain self-supervision. We extend UDALM \cite{karouzos-etal-2021-udalm}, that has shown promise for NLP tasks, for adaptation of wav2vec2 based acoustic models, and specifically XLSR. We focus on the problem of UDA in the context of a low-resource language, i.e., Greek. The key finding of our exploration is that straight-forward extension of UDALM, i.e., by using only target domain self-supervision, underperforms in this setting, and use of both source and target domain data is essential for successful adaptation. In this section, first, we will present a quick overview of the XLSR-53 training procedure, and then we are going to outline the proposed domain adaptation approach, which is shown in Fig.~\ref{fig:m2ds2}.


\subsection{XLSR-53}

XLSR-53 \cite{conneau21_interspeech} is a massively pre-trained speech model, trained on $56,000$ hours of multilingual speech, covering $53$ languages. The model is based on wav2vec2 \cite{baevski2020wav2vec}, which is composed of a multi-layer convolutional feature encoder, that extracts audio features $z_t$ from the raw audio, and a transformer context encoder that maps the latent audio features to the output hidden states $c_t$.
Each latent feature $z_t$ corresponds to $25$ ms of audio with stride $20$ ms.
A contrastive objective $L_c$ is used for pre-training. For this, product quantization \cite{jegou2010product} is applied to the features $z_t$, and then a discrete approximation of $z_t$ is obtained by sampling from a Gumbel-softmax distribution \cite{gambel_soft}, to obtain discrete code vectors $q_t$, organized into $G=2$ codebooks with $V=320$ vocabulary entries each. The contrastive loss aims to identify the correct code vector for a given time step, among a set of distractors $Q_t$, obtained through negative sampling from other timesteps. To avoid mode collapse, a diversity loss $L_d$ is included by maximizing the entropy over the averaged softmax distribution over the code vector entries $\bar{p}_g$. The total loss is:

\begin{equation}
\label{eq:xlsr}
    L_s = \underbrace{
        \textstyle - log \frac{e^{s(z_t, q_t)}}{\sum_{\tilde{q} \sim Q_t} e ^{s(z_t, \tilde{q})}}
    }_{\mathclap{\text{Contrastive Loss}} } \overbrace{
        - \frac{1}{GV}\sum_{g=1}^{G}\sum_{v=1}^V \bar{p}_{g,v}log(\bar{p}_{g,v})
    }^{\mathclap{\text{Diversity Loss}}}
\end{equation}

\subsection{Domain Adaptive finetuning for Contrastive Learning of Speech Representations}
\label{sec:m2ds2-approach}

Fig.~\ref{fig:m2ds2} shows the proposed finetuning process. The key intuition is that we want the model to synergistically learn the task at hand (in our case ASR), while being adapted to the target domain by in-domain self-supervision.
In the left we see the general pre-training stage of XLSR-53, which is pre-trained on $56$K hours of multilingual audio corpora using the contrastive pre-training objective. In the right we see the proposed finetuning stage, which is inspired by ~\cite{karouzos-etal-2021-udalm}.

During finetuning we form a mixed objective function:

\begin{equation}
\label{eq:m2ds2-loss}
L = L_{CTC}(x_s, y_s) + \alpha L_{s}(x_s) +  \beta L_{s}(x_t),
\end{equation}

where $(x_s, y_s) \sim \mathcal{S}(x,y)$, $x_t \sim \mathcal{T}(x)$, $L_{CTC}$ is the CTC objective function, optimized using transcribed source domain data, and $L_{s}$ is the contrastive loss from Eq.~\eqref{eq:xlsr}. We scale the contribution of each term using hyper-parameters $\alpha$ and $\beta$.

Note that contrary to \cite{karouzos-etal-2021-udalm}, who use only in-domain self-supervision, we leverage both source and target domain samples for the mixed self-supervision. We find that this is essential in our case to avoid mode collapse, i.e., the model using only a few of the available discrete code vectors. Simultaneous self-supervision on both the source and target data alleviates mode collapse by anchoring the target code vector space to have a similar structure as the source code vectors.

Hence we refer to this approach as Mixed Multi-Domain Self-Supervision (M2DS2).

\begin{table*}[!t]
\caption{\label{tab:grec-md}The GREC-MD corpus. We can see the duration of each split in \texttt{hours:minutes:seconds} format, as well as the number of speakers for each of the sub-corpora.}
\centering
\begin{tabular}{ccrrrrr}
\hline
\textbf{Dataset} & \textbf{Domain}  & \multicolumn{1}{c}{\textbf{Speakers}} &  \multicolumn{1}{c}{\textbf{Train}} & \multicolumn{1}{c}{\textbf{Dev}} & \multicolumn{1}{c}{\textbf{Test}}  & \multicolumn{1}{c}{\textbf{Total Duration}} \\ \hline
HParl & Public (political) speech & $387$ & \texttt{99:31:41} & \texttt{9:03:33} & \texttt{11:12:28} & \texttt{119:47:42} \\
CV & Crowd-sourced speech & $325$ & \texttt{12:16:17} & \texttt{1:57:44} & \texttt{1:59:19} & \texttt{16:13:20} \\
Logotypografia & News casts & $125$ & \texttt{51:58:45} & \texttt{9:08:35} & \texttt{8:59:22} & \texttt{70:06:42} \\
Total & - & $713$ & \texttt{163:46:43} & \texttt{20:09:52} & \texttt{22:11:44} & \texttt{206:08:19} \\ \hline
\end{tabular}
\end{table*}
\section{The Grec-md corpus}
\label{sec:grec-md-corpus}

For our experiments we compose a speech corpus for the Greek language, that is suitable for multi- and cross-domain evaluation. The GREC-MD corpus contains $206$ hours of Greek speech. Audio is segmented into individual utterances and each utterance is paired with its corresponding transcription. Table~\ref{tab:grec-md} summarizes the included sub-corpora, as well as the train, development and test splits.
The dataset is constructed with three core principles in mind:

\begin{enumerate}
    \item \textbf{Data Volume}: We collect the largest publicly available speech recognition corpus for the Greek language, able to scale to hundreds of hours of transcribed audio.
    \item \textbf{Temporal Relevance}: Language changes over time. We aim at an up-to-date corpus that encompasses the latest terms and topics that appear in daily speech.
    \item \textbf{Multi-Domain Evaluation:} Single domain evaluation can lead to misleading estimations of the expected performance for ASR models. For example, state-of-the-art ASR models \cite{gulati2020conformer} achieve under $5\%$ Word Error Rate (WER) on Librispeech  \cite{panayotov2015librispeech} test sets, but this is an over-estimation of system performance in the field. This is extenuated when considering different acoustic conditions or terminology. We consider multi-domain evaluation essential when developing and deploying real-world ASR models.
\end{enumerate}

To satisfy the first two points, we collect data from a public, continuously updated resource, i.e., the Hellenic Parliament Proceedings, where recordings of the parliamentary sessions are regularly uploaded. The benefit of using this resource is the straight-forward collection of a continuously growing, multi-speaker corpus of transcribed audio that is always up-to-date, as the parliamentary discussions revolve around current affairs. We refer to this corpus as HParl.
For the multi-domain evaluation, we merge HParl with two publicly available corpora, that have different acoustic and language characteristics. We refer to the merged, multi-domain corpus as GREC-MD.
In this Section, we will describe the collection and curation process of HParl, and present the relevant statistics for the experiments.

\begin{table}[ht]
\caption{\label{tab:proceedings-table}Plenary sessions included in HParl. The Hours column refers to the raw (unsegmented) hours of collected audio.}
\centering
\begin{tabular}{cccr}
\toprule
\textbf{Start date} & \textbf{End date} & \textbf{\#Sessions} & \multicolumn{1}{c}{\textbf{Hours}} \\ \midrule
15-02-2022 & 01-03-2022 & $10$ & $55$ \\
18-01-2019 & 01-02-2019 & $10$ & $52$ \\
28-03-2019 & 10-05-2019 & $20$ & $108$ \\
10-12-2018 & 21-12-2018 & $10$ & $88$ \\ \bottomrule
\end{tabular}
\end{table}


\subsection{Collection and Curation of HParl}
\label{sec:hparl-collection}

Modern technological advances allow for more direct government transparency, through the commodification of storage and internet speeds. In this spirit, the records of plenary sessions of the Hellenic Parliament are made publicly available, for direct access through a webpage\footnote{\url{https://www.hellenicparliament.gr/en/}}. The available video recordings date back to 2015.
For each plenary session, a video recording is uploaded, along with a full transcription that is recorded verbatim, and in real time by the parliament secretaries.
For the creation of HParl, we build a web-crawler that can traverse and download the video recordings, along with the transcriptions from the official website.
The collection process is parallelized over multiple threads, and parameterized by a range of dates and, optionally, a target corpus size in GB or in hours. For this version of HParl, we collect the plenary sessions in four date ranges, as described in Table~\ref{tab:proceedings-table}. The majority of the collected sessions are from 2019, but we also include sessions from 2018 and 2022 to include coverage of different topics.
The individual components of the HParl curation pipeline are: Audio Pre-processing, Text Pre-processing, Alignment, Post-processing, and dataset Splitting.

\begin{figure}
    \centering
    \includegraphics[width=.7\columnwidth]{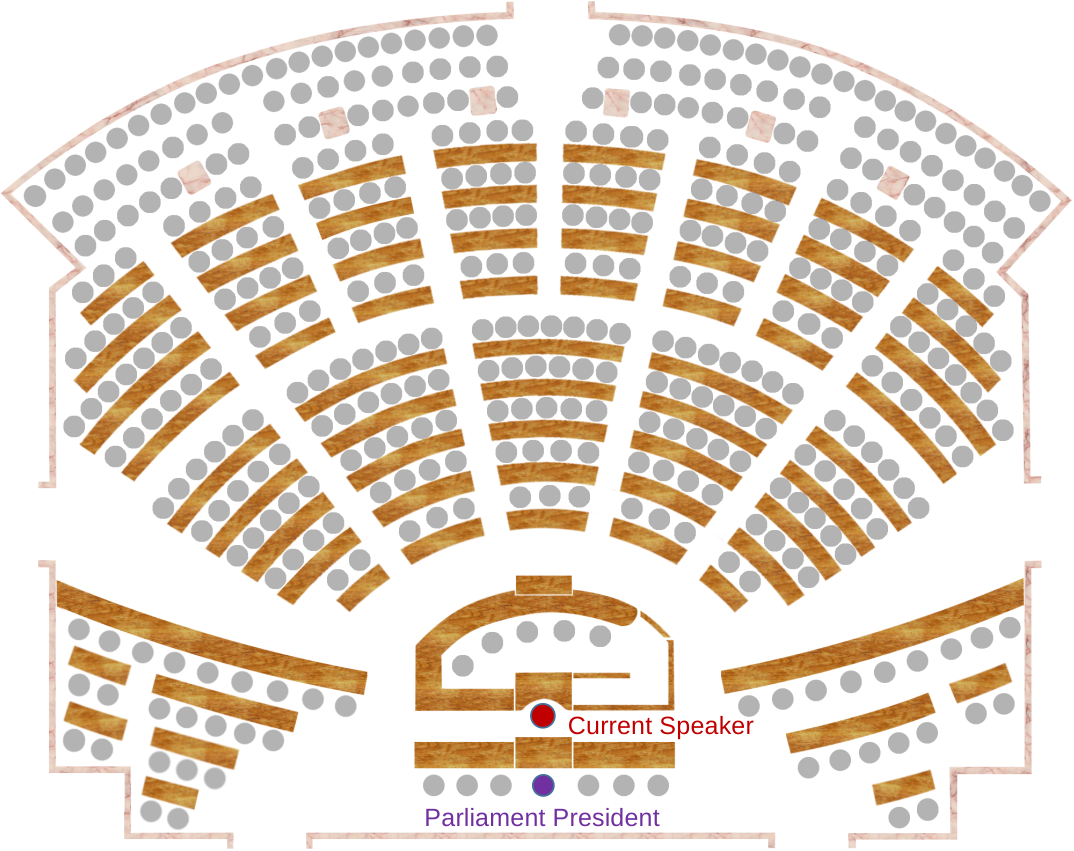}
    \caption{Overview of the Hellenic Parliament Chamber. The chamber has an amphitheatrical shape and can accomodate approximately $400-450$ people. The positions of the key speakers, i.e., current speaker and the parliament president are annotated in the image.}
    \label{fig:parliament-chamber}
\end{figure}

\subsubsection{Audio Pre-processing}
Fig.~\ref{fig:parliament-chamber} shows the layout of the Hellenic Parliament Chamber. Plenary sessions mainly take place in this room, or in the secondary House Chamber that has similar setup but is smaller in size. Because of the room and microphone characteristics, the captured audio in the video streams contains reverberation, due to sound reflections. We employ a light preprocessing pipeline, by passing the input video streams through FFmpeg, and converting them to monophonic, lossless audio format at $16000$ Hz sampling rate. The resulting audio is not passed through any de-reverberation or speech enhancement software. The resulting audio files have a minimum, average and maximum duration of $6$ minutes, $6$ hours and $16$ hours respectively.

\subsubsection{Text Pre-processing}
The text files contain full, word-by-word transcription of the speeches and questions asked by members of the audience, as well as extra annotations made by the parliament secretaries. Some annotations are relevant, i.e., the speaker name, while others are plain text descriptions of events happening during the session and need to be filtered out (e.g., ``The session is interrupted for a 15 minute break''). We use a rule-based system, based on regular expressions, that filters the unnecessary information, keeping only the transcriptions and the speaker names. The speaker labels are created by transliterating their names and roles from Greek to Greeklish using the ``All Greek to Me!'' tool \cite{chalamandaris2006all}. Text is lower-cased and normalized to remove multiple whitespaces.  The result is a text file containing the raw transcriptions, and a mapping from speaker labels to their respective text parts.

\subsubsection{Aligment and Segmentation} The primary challenge of exploiting the plenary sessions for ASR purposes is
the length of the plenary recordings, as their durations vary from $6$ minutes to $16$ hours in length.
However, data samples used to train ASR are generally less than $30$ seconds long. Computational challenges have limited the length of training utterances for HMM-GMM models \cite{MEYER2006532},  and continue to do so in the contemporary
neural network models. Therefore, we need to segment the sessions
into smaller pieces more suitable for ASR training.
A second challenge is posed by mismatches between audio and transcripts. Parliamentary proceedings do not fully capture everything that is said during the parliamentary sessions, and do not account for speech disfluencies.

In order to obtain smaller, clean segments, that are suitable for ASR training we follow the segmentation procedure proposed by \cite{8268956}. Initially the raw recordings are segmented into ~$30$ second segments and the transcriptions are split into smaller segments of approximately $1000$ words called \emph{documents}.
Each segment is decoded using a seed acoustic model trained on the Logotypografia corpus \cite{digalakis03_eurospeech} and a 4-gram biased LM trained on the corresponding transcription of each recording.
The best path transcript of each segment is obtained and paired with the best matching \emph{document} via TF-IDF similarity. Finally each hypothesis is aligned with the transcription using Smith-Waterman alignment \cite{SMITH1981195} to select the best matching sub-sequence of words. The above method yields a list of text utterances, with their corresponding start and end times in the source audio files. The procedure yields $120$ hours of useable segmented utterances out of the original $303$ hours of raw audio, or a ratio of $39.6\%$.

\subsubsection{Post-processing} After the segments are extracted, we filter out extremely short segments (less than $2$ words). Moreover, the iterative alignment algorithm may replace some intermediate words with a \texttt{<spoken-noise>} tag. When this tag is inserted, we match the surrounding text with the raw transcriptions and re-insert the missing words. Furthermore, we match each segment to its corresponding speaker label. Segments without a speaker label are discarded. Lastly, speakers are associated to their gender based on name suffixes, using a simple, Greek language-specific, rule: Speaker names which end in a(\textalpha), h(\texteta), w(\textomega) or is(\textiota\textvarsigma) are classified as female, while the rest as male. We format the segments, speaker and gender mappings in the standard folder structure used by the Kaldi speech recognition toolkit \cite{Povey_ASRU2011}.

\subsubsection{Data Splitting} We provide an official train - development - test split. The development set contains $3$ plenary sessions, one from 2018, one from 2019 and one from 2022, resulting to $9$ hours of segmented speech. Similarly, the test set contains one session from each year, resulting to $11$ hours of segmented speech. The rest $99$ hours of segmented speech are assigned to the training set.

\subsection{Including corpora from different domains}

We merge HParl with two publicly available corpora to create GREC-MD for multi-domain evaluation.

\subsubsection{Common Voice} Common Voice (CV) \cite{ardila2020common} is a crowd-sourced, multi-lingual corpus of dictated speech, created by Mozilla. The data collection is performed by use of a web app or an iPhone app. Contributors are presented with a prompt and are asked to read it. The prompts are taken from public domain sources, i.e., books, wikipedia, user submitted prompts and other public corpora. The maximum prompt length is $15$ words. A rating system is built into the platform, where contributors can upvote or downvote submitted \texttt{<audio,transcript>} pairs. A pair is considered valid, if it receives two upvotes. Speaker independent train, development and test splits are provided.
The dataset is open to the research community, released under a permisFsive Creative Commons license (CC0).
In this work, we use version 9.0 of CV, accessed on April 27, 2022. We keep only the valid utterances, i.e., $16$ hours of speech from $325$ contributors ($19-49$ years old, $67\%$ male / $23\%$ female).

\subsubsection{Logotypografia} Logotypografia \cite{digalakis03_eurospeech} is one of the first corpora for Large Vocabulary Continuous Speech Recognition in Greek. The dataset contains $33,136$ newscast utterances, or $72$ hours of speech. The utterances were collected from $125$ speakers ($55$ male, $70$ female), who were staff of the popular ``Eleftherotypia'' newspaper in Greece, under varied acoustic conditions. Approximately one third of the utterances were collected in a sound proof room, one third in a quiet room and the last third in an office room. The average utterance duration is $7.8$ seconds. The transcriptions contain several speech and non-speech events (e.g., \texttt{<cough>}), lower-cased Greek words and stress marks. Numbers are expanded to full words. We use the whole dataset, and perform light preprocessing in the transcriptions, by discarding the annotated events and punctuation.

We hence refer to each dataset by the abbreviations: HParl: HP, CommonVoice: CV, Logotypografia: LG.

\section{Experimental Settings}
\label{sec:settings}
For our experiments we use the following hyper-parameter settings, unless explicitly stated otherwise. For model training, we use AdamW optimizer \cite{loshchilov2018decoupled} with learning rate $0.0003$. We apply warmup for the first $10\%$ of the maximum training steps, and a linear learning rate decay after that. Models are finetuned for a maximum of $10000$ steps. For speech recognition training, we make use of the Connectionist Temporal Classification (CTC) loss \cite{graves2006connectionist}, optimized using the available transcribed data in each scenario. Validation runs every $500$ steps on the development set, and early stopping is employed on the development CTC loss with patience $5$. Batch size is set to $8$ during finetuning for all scenarios, except for M2DS2. In the case of M2DS2 we create mixed batches of size $12$, containing $4$ transcribed source domain samples and $8$ unlabeled target domain samples and train for $10,000$ CTC updates. For memory reasons we split the mixed batches in mini-batches of $4$ and interleave them during model training. Gradients are accumulated over $3$ interleaved batches. For the self-supervised objective, we create masks of maximum timestep length $10$, with masking probability $0.4$. We weigh the contributions of the source and target domain contrastive objectives, and bring them to the same order of magnitude as the CTC loss, by setting $\alpha=0.01$ and $\beta=0.02$.
The convolutional feature encoder is kept frozen for all experiments. Our code is based on the huggingface \footnote{\url{https://huggingface.co/docs/transformers/}} implementation of XLSR. For all experiments we resample the audio files to $16$ kHz and downsample to single channel audio. We exclude utterances in the training set that are longer than $12$ seconds. All experiments are run on a single NVIDIA RTX 3090 GPU, with mixed precision training.

For the Language model training, we create a large corpus for the Greek language using a subset of the Greek part of CCNet \cite{wenzek2020ccnet} (approximately $11$ billion tokens) and combine it with $1.5$ billion tokens from the Greek version of Wikipedia and the Hellenic National Corpus (HNC) \cite{hatzigeorgiu2000design}.
During preprocessing, we remove all punctuation and accents, deduplicate lines and convert all letters to lowercase.
We will refer to this corpus as the Generic Greek Corpus (GGC).
We train a 4-gram language model on GGC using KenLM \cite{heafield2011kenlm} and prune bigrams, trigrams and four-grams with counts less than $3$, $5$ and $7$ respectively. We incorporate the n-gram LMs at inference time using the pyctcdecode framework\footnote{\url{https://github.com/kensho-technologies/pyctcdecode}}. We use language model rescoring over a beam search decoder with $13$ beams.

The evaluation metric is the Word Error Rate (WER) over the target test set. For assessing the adaptation effectiveness we also report the relative WER improvement over the unadapted baseline in appropriate scenarios, which is defined in Eq.~\eqref{eq:rar}. We refer to this metric as Relative Adaptation Improvement (RAI) for the rest of this paper:

\begin{equation}
    \label{eq:rar}
    RAI = - \frac{WER_{adapted} - WER_{unadapted}}{WER_{unadapted}} \times 100 \%
\end{equation}

The minus sign is included, so that RAI takes negative values when the adaptation fails, i.e., when $WER_{unadapted} < WER_{adapted}$.

\begin{table}[htb!]
    \centering
    \caption{\label{tab:supervised}ASR performance of XLSR-53 over the three corpora for fully supervised in-domain finetuing (WER)}

    \begin{tabular}{c|cc}
\toprule
    \diagbox{Dataset}{LM}     & No LM & 4g GGC \\
\midrule
        HP & $26.21$& $15.64$ \\
        CV & $29.33$ & $9.52$\\
        LG & $31.94$& $26.45$\\
    \bottomrule
    \end{tabular}
\end{table}
\section{Supervised In-Domain Training}
\label{sec:supervised}

In the first set of experiments, we explore the performance of supervised  finetuning of XLSR-53 for each domain. This will give an  upper bound estimation for UDA performance. We finetune XLSR-53 on CV, HP and LG (separately) and perform in-domain evaluation on the respective test sets. Results are summarized in Table~\ref{tab:supervised}. The first row indicates the performance of greedy decoding, while in the second row we report the performance of the beam search decoder, rescored using the scores of the 4-gram GGC language model. We observe that the greedy decoding performance is under $30$ WER for both HP and CV, while for LG we achieve $\sim 32$ WER. This makes sense, as LG is the most diverse dataset, with respect to the included acoustic conditions. Furthermore, we observe that the incorporation of a language model results in an impressive WER reduction on CV, followed by HP and then LG. While CV includes relatively simple phrases with common vocabulary, HP and LG contain more specialized terminology.

\begin{table*}
\centering

\caption{\label{tab:results}M2DS2 performance using greedy decoding for UDA between HP, CV, and LG. A  $\rightarrow$ B indicates that A is the source domain and B is the target domain. (G) indicates greedy decoding. (LM) indicates beam search with LM rescoring. We report the WER on the target test set, as well as the RAI (\%) over the SO (unadapted) baseline. WER: lower is better. RAI: higher is better.}
  \resizebox{\textwidth}{!}{%

\begin{tabular}{l| c | cr | cr | cr  || c | cr | cr | cr}
\toprule
  Method  & SO (G)& \multicolumn{2}{c|}{CPT (G)}  &\multicolumn{2}{c|}{PSL (G)} & \multicolumn{2}{c||}{M2DS2 (G)}     & SO (LM) & \multicolumn{2}{c|}{CPT (LM)}  &\multicolumn{2}{c|}{PSL (LM)} & \multicolumn{2}{c}{M2DS2 (LM)} \\
  Setting  & WER & WER & RAI & WER & RAI &WER & RAI    & WER & WER & RAI & WER & RAI & WER & RAI\\
\midrule
    HP $\rightarrow$ CV & $55.9$ & $59.68$ & $-6.8$ & $55.3$ & $1.2$ & $\mathbf{52.95}$ & $\mathbf{5.3}$
    & $25.26$& $26.44$ &$-4.7$ & $24.24$ & $4.0$& $\mathbf{18.35}$ & $\mathbf{27.4}$\\
    HP $\rightarrow$ LG & $\mathbf{48.65}$ & $52.63$ & $-8.2$ & $57.68$ & $-18.6$ & $58.99$ & $-21.3$
    & $\mathbf{30.34}$ & $32.27$ & $-6.4$&$39.32$ & $-29.6$ &$32.58$ & $-7.4$\\
    LG $\rightarrow$ CV & $59.57$ & $66.43$ & $-13.4$ & $81.90$ & $-39.8$ & $\mathbf{51.31}$ & $\mathbf{12.4}$
    & $25.96$&$31.51$ & $-21.4$ &$52.05$ &$-100.5$ &$\mathbf{17.30}$ &$\mathbf{33.4}$\\
    LG $\rightarrow$ HP & $62.13$ & $67.51$ & $-8.7$ & $71.46$ & $-15.0$ & $\mathbf{60.09}$ & $\mathbf{3.3}$
    &$31.48$ &$31.58$ &$-0.3$ &$45.36$ &$-44.1$ &$\mathbf{31.36}$ &$\mathbf{0.4}$\\
    CV $\rightarrow$ LG & $69.55$ & $71.12$ & $-2.3$ & $71.34$ & $-2.6$ & $\mathbf{63.40}$ & $\mathbf{8.8}$
    &$50.80$ &$52.40$ &$-3.2$ &$48.68$ &$4.2$ &$\mathbf{36.93}$ &$\mathbf{27.3}$  \\
    CV $\rightarrow$ HP & $70.72$ & $73.83$ & $-4.4$ & $78.05$ & $-10.4$ & $\mathbf{68.70}$ & $\mathbf{2.9}$
        &$52.09$ & $52.18$& $-0.2$& $54.82$& $-5.2$&$\mathbf{41.88}$ &$\mathbf{19.6}$\\

\bottomrule
\end{tabular}
}

\end{table*}
\section{Unsupervised Domain Adaptation Using In-Domain Audio}
\label{sec:experiments}

Here, we evaluate the effectiveness of M2DS2 for UDA. We compare with three baselines:
\begin{enumerate}
    \item \textbf{Source Only Training (SO):} We perform supervised finetuning of XLSR-53 (CTC) using only the source-domain data, and run decoding on the target domain test set. No in-domain data are used for adaptation.
    \item \textbf{Continual Pre-Training (CPT):} We perform a pre-training phase using the loss in Eq.~\eqref{eq:xlsr} on the target domain train set, to create adapted versions of XLSR. Pre-training is run for $20000$ steps with batch size $4$. Only the audio is used, without transcriptions. The adapted checkpoints are then finetuned by use of CTC loss on the source domain transcribed data. Evaluation is performed on the target test set.
    \item \textbf{Pseudolabeling (PSL):} We finetune XLSR-53 using the source domain data with CTC loss. Then we run inference on the source model, to extract silver transcriptions for the target domain training set. We use the silver transcriptions for supervised finetuning on the target domain.
\end{enumerate}

In Table~\ref{tab:results} we compare M2DS2 with the SO, CPT and PSL baselines for six adaptation scenarios, i.e., cross dataset evaluation between the three datasets in GREC-MD. The left half corresponds to greedy decoding, while for the right half we use the 4-gram LM trained on GGC.
First, we observe the SO model performance. The SO models are the finetuned models from Table~\ref{tab:supervised}, evaluated in out-of-domain settings. We see that out-of-domain evaluation results in a large performance hit, e.g., while in the CV9 $\rightarrow$ CV9 in-domain setting we achieve $29.33$ WER,  in the CV9  $\rightarrow$ HP out-of-domain setting we get $69.55$ WER. This confirms that for real-world ASR tasks, multi-domain evaluation is of essence. Second, we observe that in most adaptation scenarios both CPT and PSL fail to surpass the SO (unadapted) baseline. In the case of CPT, we hypothesize that is due to the relatively data constrained version of our setting. In the best-case scenario, we have $99$ hours of available target domain audio, which is not enough to perform a discrete CPT stage. Note that most of works in the literature use $\sim 1000$ hours of target audio for CPT. In the case of PSL, the poor performance is due to the quality of the silver labels created by the seed model. While the performance would improve with more elaborate approaches (e.g., confidence filtering), in challenging adaptation scenarios PSL approaches are limited by the SO model's performance.
Lastly, we observe that M2DS2 is the only approach among our baselines that manages to achieve a positive RAI in most adaptation scenarios, by consistently outperforming the SO baseline by significant margins. This is exaggerated when we include a LM during inference.
One exception in this pattern is the HP $\rightarrow$ LG scenario, where the SO baseline achieves the best performance. We attribute this to the fact that we performed minimal hyper-parameter tuning during model development.

\begin{figure}
    \centering
    \includegraphics[width=.9\columnwidth]{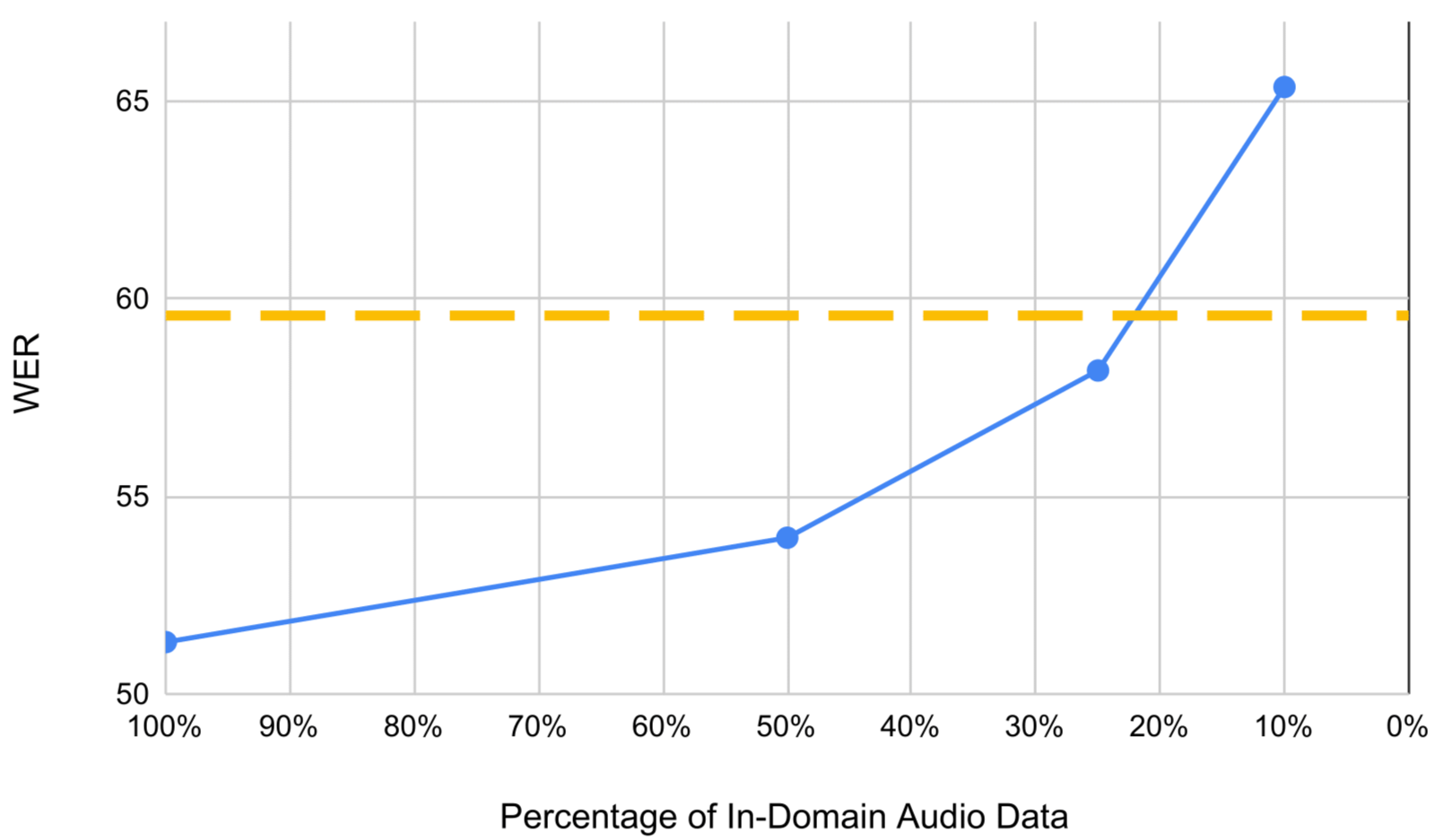}
    \caption{Performance of M2DS2 (blue line) for the LG $\rightarrow$ CV setting, when reducing the amount of available target samples to $50\%$, $25\%$, and $10\%$ of the original dataset (horizontal axis). SO performance is indicated with the orange line. Vertical axis: WER, Horizontal Axis: target audio percentage ($100\%\rightarrow 0\%$) }
    \label{fig:sample-eff}
\end{figure}
\subsection{The sample efficiency of M2DS2}
\label{sec:sample-eff}

One key observation in the literature, and in our experiments is that CPT requires a large amount of un-transcribed target domain audio. This raises the question, can we leverage self-supervision for domain adaptation in data constrained settings?

In Fig.~\ref{fig:sample-eff} we evaluate the performance of M2DS2, when we reduce the amount of target domain audio. Specifically we focus on the scenario of LG $\rightarrow$ CV. The full training corpus of CV contains $12$ hours of audio. We train M2DS2 with $50\%$, $25\%$ and $10\%$ of the available samples, or $6$, $3$ and $1.2$ hours of audio respectively, and plot the resulting WER on the target (CV) test set. In all cases, the full source (LG) training corpus is used.
We observe that M2DS2 achieves lower WER than the SO baseline, even with only $3$  hours of target domain audio. While CPT can suffer from catastrophic forgetting, as most multi-stage training approaches, M2DS2 avoids this issue, being a single-stage approach with a mixed task-specific and self-supervised objective.
This provides a promising avenue for adaptation, when collection of in-domain recordings is expensive, or infeasible.





\begin{figure}[ht]
\centering
\subfloat[Only target domain self-supervision]{%
\includegraphics[width=.9\columnwidth]{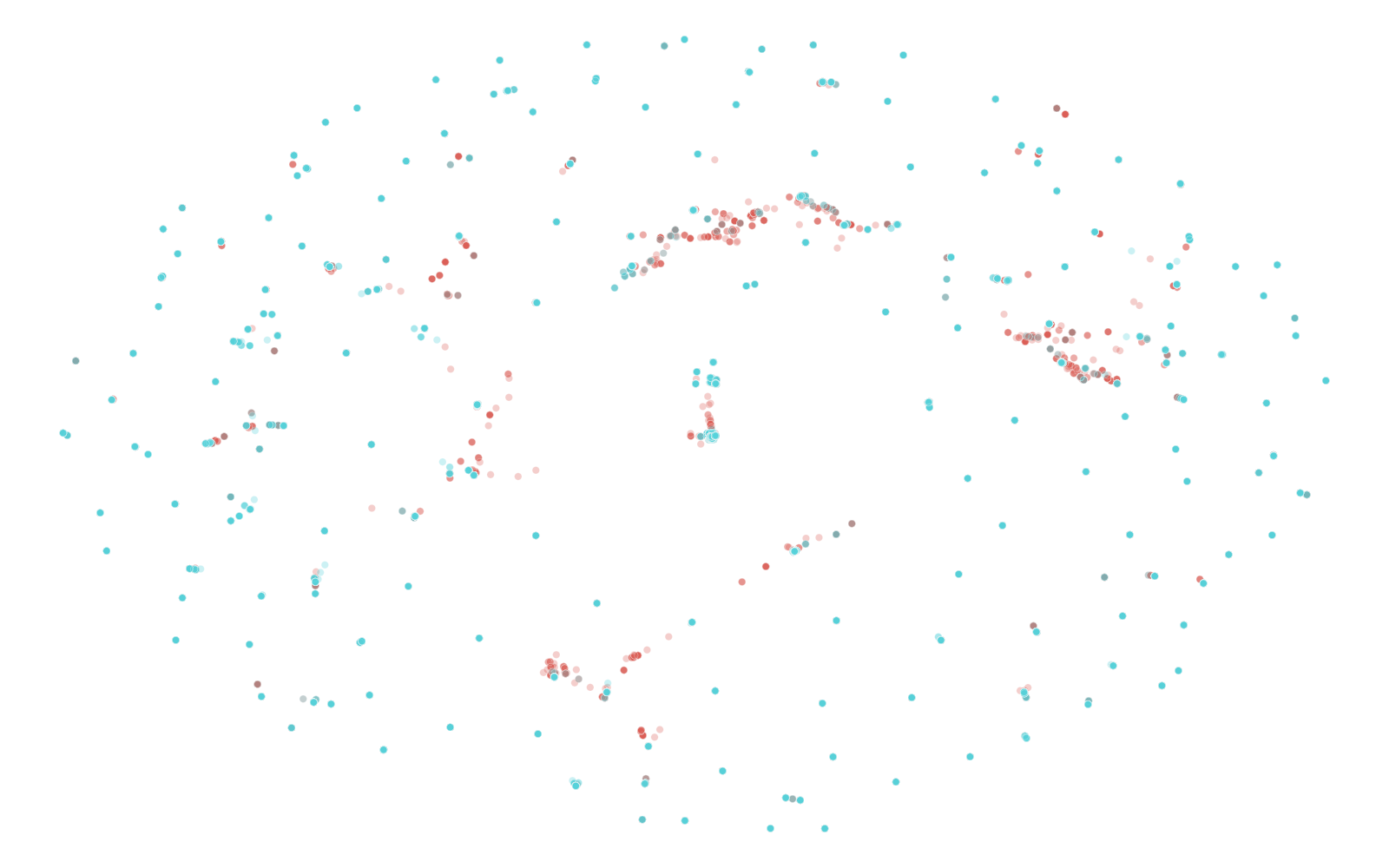}%
}\quad
\subfloat[Target and source domain self-supervision]{%
  \includegraphics[width=.9\columnwidth]{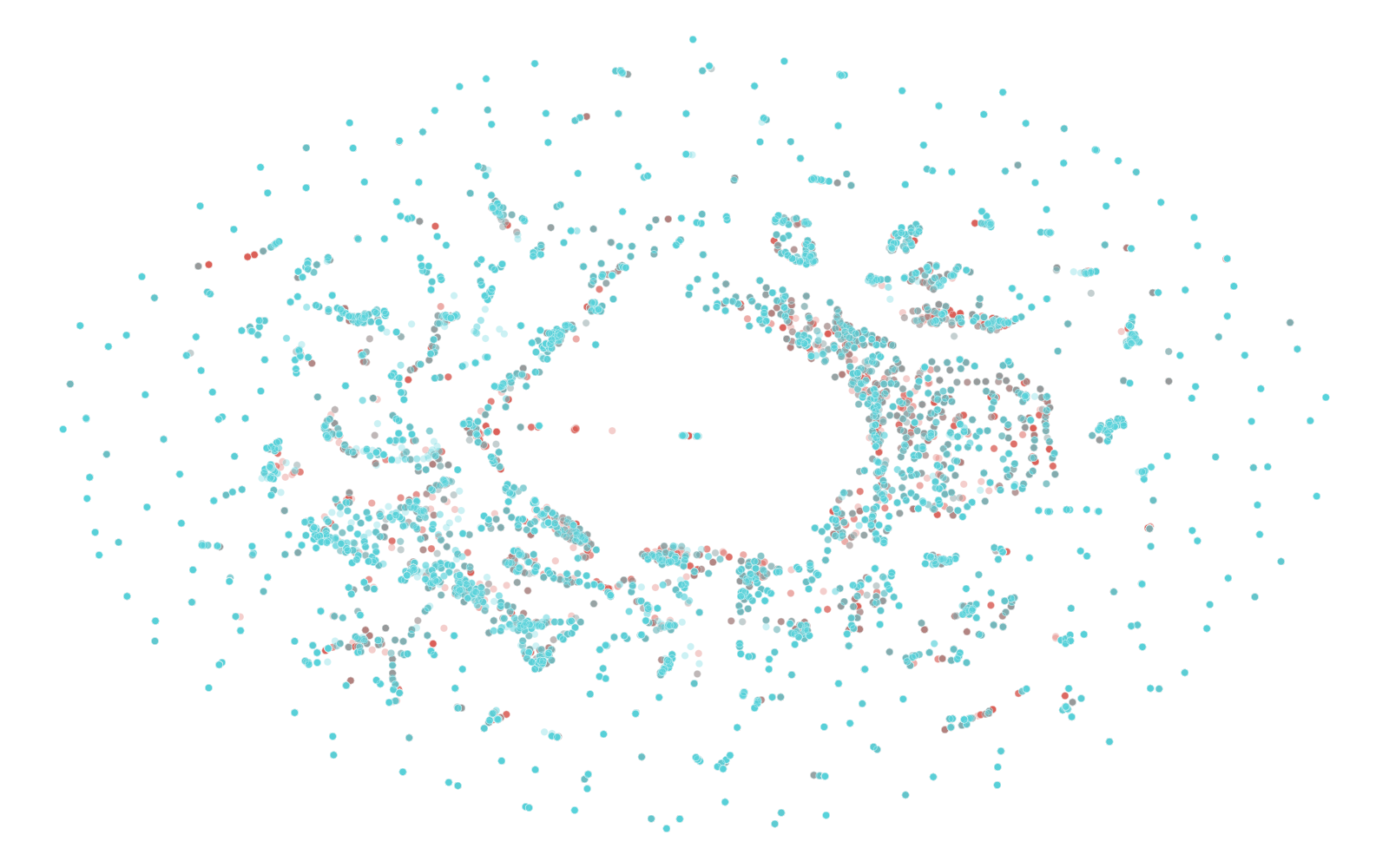}
}
\caption{T-SNE scatter plots of code vectors extracted from M2DS2 without source domain self-supervision (top) and with source domain self-supervision (bottom) for LG (red) and CV (teal)}
\label{fig:tsne}
\end{figure}

\subsection{The importance of Multi-Domain Self-Supervision}
\label{sec:multi-domain}

In Section~\ref{sec:m2ds2-approach} we argue that it is essential to include both source and target domain data for the self-supervised objective of M2DS2. To illustrate the effect of this approach, we train two versions of M2DS2 for the LG $\rightarrow$ CV scenario. For the first version we set $\alpha=0.01$, while for the second we set $\alpha=0$, removing the second term of Eq.~\eqref{eq:m2ds2-loss}. We extract the code vectors for the first $100$ samples of both LG and CV, and flatten them across the time steps , resulting to $60000 \times 768$ code vectors corresponding to individual timesteps. We plot these code vectors using T-SNE \cite{van2008visualizing} in Fig.~\ref{fig:tsne} for both models. We see that when we do not include the source domain self-supervision, the code vector space collapses in a few tight clusters, and most audio segments correspond to just a few code vectors. This is a visual clue that indicates the mode collapse problem. When we include the source domain term, we see that the that the code vector space has more structure, and coverage of the space is more complete, both for CV (target domain) and LG (source domain). Experimentally we train M2DS2 with $\alpha=0$ for all source / target domain pairs and we find that the mode collapse is destructive for target domain performance. During our experiments we got WER in the range $80 - 99$, indicating failure to converge to acceptable solutions across all scenarios. The simple inclusion of both source and target domain self supervision stabilizes training, avoids mode collapse and leads to successful unsupervised adaptation between domains.

\begin{table}
    \centering
        \caption{\label{tab:m2ds2-lm}Language adaptation of the M2DS2 LG $\rightarrow$ CV model, using biased and augmented LMs. We use the variant of the model trained with $3$ hours of in-domain audio. We vary the amount of in-domain text data from $752$K tokens to $38$K tokens.}
    \begin{tabular}{c|cc}
    \toprule
         & Biased LM & Augmented LM  \\
        \midrule
        $100\%$ & $11.22$ & $12.84$ \\
         $50\%$	& $15.13$ & $15.05$ \\
        $25\%$ &	$20.84$ &	$16.64$ \\
        $10\%$ &	$27.75$	& $18.47$ \\
        $5\%$	& $33.04$ &	$19.31$ \\
        \midrule
        Baseline (M2DS2 + Generic LM) & \multicolumn{2}{c}{$20.7$}\\
        \bottomrule
    \end{tabular}

\end{table}

\section{Unsupervised and Weakly Supervised Language Adaptation}

\label{sec:weak}

\begin{figure}
    \centering
    \includegraphics[width=.9\columnwidth]{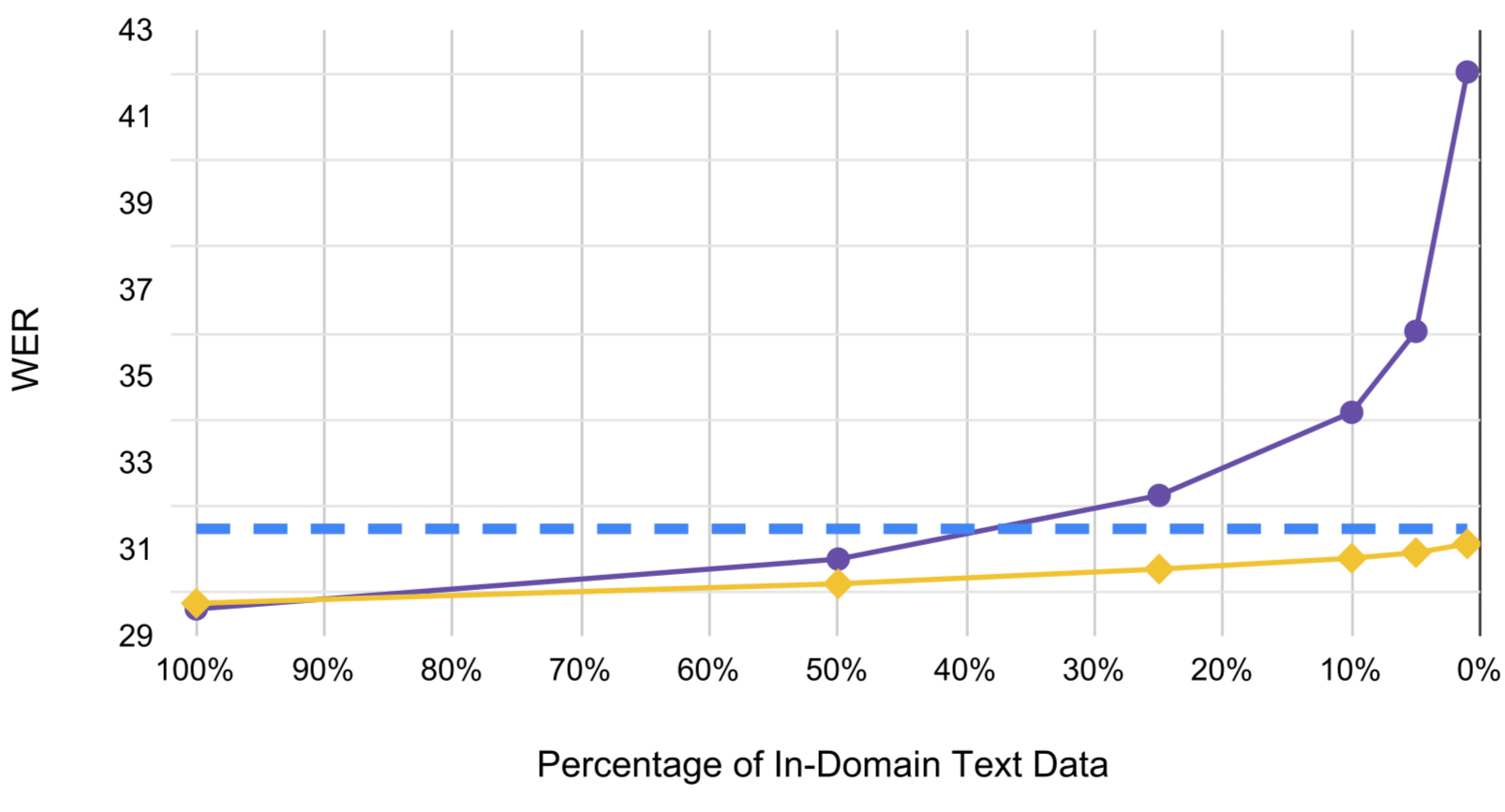}
    \caption{Language-only adaptation for LG $\rightarrow$ HP using the SO model finetuned on LG. In-domain text data range from $11$M tokens (left) to $110$K tokens (right). Blue/dashed: Baseline with generic LM. Purple/circles: Biased LM. Orange/diamonds: Augmented LM.}
    \label{fig:so-lm}
\end{figure}

When small amounts of in-domain textual data are available, simple N-gram LM adaptation techniques can be very effective. In this brief set of experiments, we first explore the unsupervised language adaptation setting, where no in-domain audio is used, and then we relax the problem to the weakly supervised setting, where M2DS2 is combined with the adapted N-Gram LMs. These settings are described in Sections~\ref{sec:unsup-lang-prob} and~\ref{sec:weak-lang-prob} respectively.
We explore two approaches for LM adaptation: biased LMs, and in-domain data augmentation. To create biased LMs, we train a 4-gram LM on the available in-domain data. Then we replace the generic LM trained on GGC.
For LM data augmentation we follow a perplexity filtering approach similar to \cite{wenzek2020ccnet}. We first train a biased LM using available target domain text, and then use it to calculate the perplexity of each line in the GGC corpus. We keep the $10\%$ of the lines with the lowest perplexity. Then we train a 4-gram LM on the augmented ``in-domain'' corpus and use it for inference.

Fig.~\ref{fig:so-lm} shows the performance of the SO LG $\rightarrow$ HP model with biased and augmented LMs, as we reduce the amount of available in-domain text data from $100\%$ to $1\%$ of the in-domain transcriptions ($11$B tokens to $110$K tokens respectively).
As a baseline we include the LG $\rightarrow$ HP SO model in combination with the generic LM trained on GGC.
We observe that the use of biased LMs can lead to successful adaptation, when an adequate amount of in-domain text data is available. On the other hand the LM augmentation approach results to successful augmentation, even with very small amounts of in-domain text.

In Table~\ref{tab:m2ds2-lm} we see the results of LM adaptation, combined with the M2DS2 LG $\rightarrow$ CV model. To demonstrate the sample efficiency of the approach, we use the variant that was trained using only $25\%$ of the target domain audio ($3$ hours). We compare with M2DS2 combined with the 4-gram GGC LM for inference. We draw similar conclusions, i.e., use of biased LMs performs well for sufficient text data. When we use augmented LMs we can leverage very small amounts of in-domain text.

\begin{table}
    \centering
        \caption{\label{tab:closing-the-gap}Closing the gap between SO training and fully supervised training for the LG $\rightarrow$ CV adaptation scenario using M2DS2, with varying amounts of available unpaired in-domain audio and text. (U): unsupervised acoustic or language adaptation. (W): weakly supervised adaptation.}
    \begin{tabular}{c|cccc}
    \toprule
        Method & \#Audio (h) & \#Tokens &  LM  & WER \\
        \midrule
      SO (U) & - & - & N/A & $59.57$ \\
      M2DS2 (U) &   $3$	& - & N/A & $57.31$ \\
      M2DS2 (U) &   $12$	& - & N/A & $51.31$ \\
      \midrule
      SO (U) & - & - & Generic & $25.96$ \\
      SO (U) & - & $38,632$ & Augmented & $24.67$ \\
      SO (U) & - & $751,953$ & Augmented & $20.46$ \\
      M2DS2 (U) & $3$ & - & Generic & $20.7$ \\
      M2DS2 (U) & $12$ & - & Generic & $17.3$ \\
      \midrule
      M2DS2 (W) & $3$ & $38,632$ & Augmented & $19.31$ \\
      M2DS2 (W) & $12$ & $38,632$ & Augmented & $16.29$ \\
      M2DS2 (W) & $3$ & $751,953$ & Augmented & $12.84$ \\
      M2DS2 (W) & $12$ & $751,953$ & Augmented & $10.61$ \\
      \midrule
      Supervised & $12$ & $751,953$ & Generic & $9.52$ \\
      Supervised & $12$ & $751,953$ & Augmented & $7.94$ \\
        \bottomrule
    \end{tabular}

\end{table}

\section{Discussion \& Conclusions}

In this work, we have explored Unsupervised and Weakly Supervised Domain Adaptation of ASR systems in the context of an under-resourced language, i.e., Greek. We focus on domain adaptation through in-domain self-supervision for XLSR-53, a state-of-the-art multilingual ASR model. Specifically, we adopt a mixed task and self-supervised objective, inspired from NLP, and show that using only in-domain self-supervision can lead to mode collapse of the representations created by the contrastive loss of XLSR-53. Therefore, we propose the use of mixed task and multi-domain self-supervision, M2DS2, where the contrastive loss leverages both the source and target domain audio data. For evaluation we create and release HParl, the largest to-date public corpus of transcribed Greek speech ($120$ hours), collected from the Greek Parliamentary Proceedings. HParl is combined with two other popular Greek speech corpora, i.e., Logotypografia and CommonVoice, for multi-domain evaluation.

In our experiments, we find that while most UDA baselines fail in our low-resource setting, the proposed mixed task and multi-domain self-supervised finetuning strategy yields significant improvements for the majority of adaptation scenarios. Furthermore, we focus our ablations on showcasing the sample efficiency of the proposed finetuning strategy, and demonstrating the necessity of including both source and target domain data for self-supervision. Finally, we show that M2DS2 can be combined with simple language model adaptation techniques in a relaxed weakly supervised setting, where we achieve significant performance improvements with a few hours of in-domain audio and a small, unpaired in-domain text corpus.

More concretely, in Table~\ref{tab:closing-the-gap} we present a summary of the discussed  unsupervised and weakly supervised adaptation combinations, for different amounts of available in-domain audio and text. Note that for the weakly supervised scenarios, the in-domain audio and text are unpaired. We see, that when no in-domain data are available, including an n-gram LM trained on large corpora is recommended. Furthermore, when in-domain audio is available, following a mixed multi-domain finetuning strategy using M2DS2 can yield significant WER reductions, even for a few hours of audio.
When small amounts of in-domain text is available, using a corpus augmentation strategy, e.g., perplexity filtering, can produce adapted LMs and yield small improvements to the final WER.
In the case of sufficient amounts of unpaired in-domain text and audio, independent adaptation of XLSR-53 using the audio data and the n-gram LM using the text data can yield comparable performance with a fully supervised finetuning pipeline.

\section{Future Work}

In the future we plan to explore the effectiveness of the proposed adaptation strategy for other languages, and different adaptation settings, e.g., accent or cross-lingual adaptation. Of special interest is the investigation of the effectiveness of our approach for endagered languages, e.g., Pomak. Furthermore, we plan to explore the combination of in-domain self-supervision, when combined with other popular UDA techniques, e.g., teacher student models, adversarial learning, and data augmentation approaches. On the language adaptation side, we plan to explore multi-resolution learning, which has shown promise for ASR \cite{paraskevopoulos2020multimodal}, and investigate more elaborate end-to-end weakly supervised adaptation methods. Finally, we plan to expand our study in a multimodal setting, where both audio and video are available, e.g., lip reading.



\small
\bibliographystyle{IEEEbib}
\bibliography{refs}

\end{document}